# A study on performance limitations in Federated Learning

Karthik Mohan

University of Toronto
karthik.mohan@mail.utoronto.ca

**ABSTRACT**

Increasing privacy concerns and unrestricted access to data lead to the development of a novel machine learning paradigm called Federated Learning (FL). FL borrows many of the ideas from distributed machine learning, however, the challenges associated with federated learning makes it an interesting engineering problem since the models are trained on edge devices. It was introduced in 2016 by Google, and since then active research is being carried out in different areas within FL such as federated optimization algorithms, model and update compression, differential privacy, robustness, and attacks, federated GANs and privacy preserved personalization. There are many open challenges in the development of such federated machine learning systems and this project will be focusing on the communication bottleneck and data Non IID-ness, and its effect on the performance of the models. These issues are characterized on a baseline model, model performance is evaluated, and discussions are made to overcome these issues.

## 1. INTRODUCTION

Federated learning (FL) is a machine learning technique that trains an algorithm across multiple decentralized devices or servers while preserving data samples locally without exchanges to a central server. In simpler terms, it is a way of bringing machine learning models to the data rather than moving data from its source to datacenters [1]. Traditionally, machine learning models are created by running the training jobs in data centers [2]. To address the increasing concerns on data privacy, organizations are now interested in training the models on client devices with the client's data and sending back only the model parameters to aggregate them and create a final global model that can then be evaluated and deployed to everyone. There are still many open questions yet to be addressed in FL. For instance, performance optimizations are especially important for the final model convergence. In this project, I aim to study the existing issues/bottlenecks in federated learning Specifically, I will focus on the following two issues, implement a baseline FL model, and characterize the problems by presenting quantitative evidence:

1. **Communication overhead**, is one of the major bottlenecks in federated learning [3]. Since wireless links and other end-user internet connections (client devices) typically operate at lower rates than intra- or inter datacenter links and can be potentially expensive and unreliable. This communication overhead slows down the convergence of the Machine Learning algorithms. For example, the client devices could be self-driving cars in which the goal might be to create a driver sleep prevention face recognition machine learning system preventing road accidents or making use of large volumes of traffic training data from cameras in the vehicles to improve the vehicle AI agent's driving capability. Because in both cases, due to the possibility of collecting large number of samples by increasing the client devices, the data used to train models will have a large variance (carries more Information) and will be more robust to bias (race of the driver, different types of roads, and pedestrian scenarios) and thus underrepresentation of samples is minimized. The slower client connections might also cause stragglers. Hence in such non-typical client devices, it is necessary to minimize the communication overhead.

2. **Statistical Heterogeneity**, where different clients can hold vastly different amounts of data [4] or some clients will not have data for all the classes in a multiclass classification task. This violates the IID-ness (Independent and Identically Distributed) assumption - the fundamental assumption that guarantees the correctness of machine learning algorithms and statistical models. Most existing FL frameworks do not deal with this issue specifically but do allow developers to customize this process. For instance, TensorFlow Federated Learning (TFF), a popular framework developed by Google for aiding federated learning research, simply ignores the clients that do not have enough data. In this case, ignoring client data is not beneficial since the data could indeed be used to train the model by correcting for the non IID-ness. It is also well known that data is a valuable resource, and its collection is expensive. Generally, in Machine learning, the more data used to train a model, the more robust is the model's prediction thereby resulting in better model performance.

## 2. GOAL

To study the performance bottlenecks in Federated learning using the existing TensorFlow Federated framework (TFF), specifically on communication overheads and Non IID-ness in the client devices under cross-device horizontal federated learning setting. These bottlenecks are characterized on a baseline model and proposals are made to improve the model's performance with respect to each issue.

## 3. BACKGROUND

### 3.1 Existing learning settings

There are several learning techniques to train a machine learning model. The typical learning settings are:

### a. Datacenter Distributed Learning

The model is trained on a large single dataset [5, 6]. The computations are done in nodes of a single cluster or datacenter. Such learning is centrally orchestrated. It can be noted that a large amount of computation is the bottleneck in this learning setting.

### b. Cross-Silo Federated Learning

In this type of learning, the data stays at the origin since it is a type of federated learning. The clients are geo-distributed datacenters. A central orchestration server is used to aggregate, and its visibility is restricted to only the model parameters [5]. In this type of learning, the bottleneck is in communication and/or computation.

### c. Cross-Device Federated Learning

In this type of learning setting, mobile devices are the client devices. Ex: Google Keyboard. It can make suggestions and autocorrections and the underlying model is built using cross device federated learning [7].

The number of client devices in this learning setting is much larger than the above two learning settings. Depending on the task, communication could be the major bottleneck since the client devices typically use slower connections like Wi-Fi.

The clients participating in one round of computation may not participate in the next round due to reasons such as disconnectivity since the client devices are not connected to a constant power grid and most of them operate on a battery. Hence, the client devices here are stateless.

An increase in the number of devices also increases the chances of failure. Hence, the clients in this setting are highly reliable. In the interest of time, this project focuses only on this learning setting.

### d. Fully decentralized peer to peer learning

Although this type of learning setting is currently not considered, it will be included in the future scope of this project. In this type of learning, there is no central orchestration as seen in the previous learning methods. This peer-peer system enables building a trustless federated learning system, the discussion on benefits of which are currently out of scope for this project.

### 3.2 Types of cross-device federated learning.

There are three types of cross-device federated learning.

*i) Horizontal Federated Learning:*

In this type, all the client devices carry the same features.

For example, All the client devices carry the user's purchase history from store A. This type of learning accounts for homogenous learning.

*ii) Vertical Federated Learning:*

The client devices carry different features of different datasets, and this is a joint federated learning. For example, client A might have information about the user's purchase history on an online Store A and client B might have information about purchase history on offline Store B, and combined intuition from both datasets is needed to analyze the user's decision to purchase a product.

*iii) Federated Transfer Learning:*

Pretrained models are used for one task and can be utilized for another task in a federated setting. At the time of this paper, there are no concrete applications of this type of learning.

### 3.3 Frameworks for Federated Learning research

The active research interests in keeping the models training and data at the edge devices have led to the development of well-defined frameworks that supports federated learning experimentations at ease.

i) *TensorFlow Federated*[1]

Developed by Google, TensorFlow Federated (TFF) allows researchers to conduct federated learning simulations. The TFF framework provides starter federated learning datasets that are a good representative of the real-world datasets for most common machine learning like computer vision and Natural Language Processing. This project uses TFF framework.

ii) *PyTorch Mobile*[2] It is a framework developed by the PyTorch team at Facebook to execute ML models on edge

---

[1] Tensorflow Federated Learning:
  https://www.tensorflow.org/federated/get_started

[2] PyTorch Mobile: https://PyTorch.org/mobile/home/

devices for reducing latency and preserve privacy. The current version is in beta release.

iii) *Flower: A Friendly Federated Learning Framework*

It is a simplified framework that allows developers to configure a Flower server and a Flower client using one of the most popular machine learning frameworks like TensorFlow and PyTorch. This framework prioritizes the usability for developers in building federated machine learning pipelines [8]. But for experimenting, this framework does not provide a fine-grained control over the process.

### 3.4 Federated Averaging Algorithm

Deep learning training mainly relies on variants of the Stochastic gradient descent algorithm in which gradients are computed on a random subset of the dataset and these gradients are used to make one update on the parameters ie., perform one gradient descent step.

Federated SGD is a direct transposition of this algorithm to a federated setting. A more generalization of the Federated SGD algorithm is called the Federated Averaging (FedAvg) [9, 10].

**Algorithm 1** FederatedAveraging. The $K$ clients are indexed by $k$; $B$ is the local minibatch size, $E$ is the number of local epochs, and $\eta$ is the learning rate.

**Server executes:**
  initialize $w_0$
  **for** each round $t = 1, 2, \ldots$ **do**
    $m \leftarrow \max(C \cdot K, 1)$
    $S_t \leftarrow$ (random set of $m$ clients)
    **for** each client $k \in S_t$ **in parallel do**
      $w_{t+1}^k \leftarrow$ ClientUpdate$(k, w_t)$
    $w_{t+1} \leftarrow \sum_{k=1}^{K} \frac{n_k}{n} w_{t+1}^k$

**ClientUpdate**$(k, w)$:   // Run on client $k$
  $\mathcal{B} \leftarrow$ (split $\mathcal{P}_k$ into batches of size $B$)
  **for** each local epoch $i$ from $1$ to $E$ **do**
    **for** batch $b \in \mathcal{B}$ **do**
      $w \leftarrow w - \eta \nabla \ell(w; b)$
  return $w$ to server

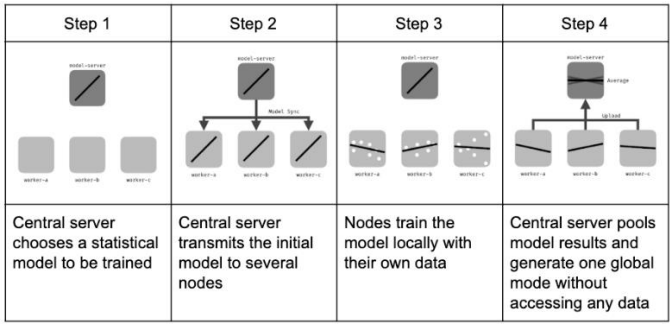

**Fig 1: Steps involved in a FedAvg Algorithm [9]**

A single round of the algorithm consists of the following steps:

1. Client Selection: The server selects the client devices that meets the specified explicit constraints. These constraints could be based on the data or the client device type.

2. Broadcast: The selected clients, get the recent version of the parameters from the server and a training program. For example, this could be a dataflow graph from the TensorFlow framework.

3. Client Computation: The clients train the downloaded model using the program and data that is available locally. This computation is a typical model training process in a client device.

4. Aggregation: The server collects the parameters from all clients that participated in the round and aggregates them. Stragglers - the devices in which the local training is much slower could be dropped.

5. Model Update: The server then updates the shared model based on the aggregated update values from the participated clients.

The communication protocol is synchronous. But recently, many new algorithms have been proposed like FedAt – Federated Asynchronous Tier [11], semi-synchronous federated averaging algorithm [12], and self-balancing Federated learning [13]. These algorithms have not yet been rigorously tested and implementation using a lower-level Federated core impacts the project duration. In the interest of time for this project, I focus only on the Federated Averaging algorithm which is supported by the TensorFlow federated framework out of the box for simulations.

### 3.5 Motivation and techniques for solving the communication bottleneck.

The client devices in the cross-device federated learning setting are typically mobile devices. Hence, the model training should happen without interfering and impacting the user's activities on the devices. For example, the training should ideally happen on the mobile devices when the user is in a meter less network like the way software updates are pushed these days. Also, typically the user's connection bandwidth is lower than a typical datacenter, which is a traditional setting to train machine learning models. This bottleneck however depends heavily on the type of machine learning task. For example, if it is a Computer Vision task, then the model parameters will be large and require more bandwidth to speed up the convergence. Since, the devices with less bandwidth act as stragglers which will then be dropped by the Federated Averaging algorithm due to timeout [14].

This limited memory, computation resources, and communication bandwidth on the client devices motivate the idea of compressing the models before sending them to the client devices (broadcasted bits) and compressing the gradients from client devices (aggregated bits) in each round [15].

The optimal tradeoff between communication and model accuracy is not yet well studied and it remains an open research problem.

The current research suggests the following to address this issue,

i) Gradient Compression – Reduce the size of objects from clients to parameter server.

ii) Model Broadcast Compression – Reduce the size of the initial model sent from the server to the clients.

iii) Local Computation reduction – Optimizing machine learning training on clients. These are general Machine learning optimizations in algorithms, training procedures and preprocessing steps done on a single device. For example., using momentum-based optimizers for training.

Gradient compression has the most impact on runtime since the clients (for ex., mobile devices) typically have slower upload connections than the download. And it is also necessary to reduce the costs of the device's user that is associated with the transfer to model parameters to the server.

Model broadcast and update compression can be done using one of the existing compression techniques such as quantization. An implementation of the model broadcast and update compression using a lossy compression algorithm is discussed briefly in the Experimentation section.

## 3.6 Motivations and techniques for solving the Non IID-ness.

Traditional Machine Learning algorithms work on the assumption that data is Identical and Independently Distributed (IID), this is one of the important statistical assumptions which guarantees the correctness of Machine Learning Algorithms.

This assumption is easily violated when the data distribution is different across many client devices. At a higher level, the non IID-ness can occur due to the following cases:

1. Data Imbalance.
2. Missing Classes.
3. Missing Features.
4. Missing Values.

One of the recent methods developed by researchers [16] to handle the Non IID-ness is using bound aware modeling in which the client device models are piece-wise averaged based on their bounds. In Fig 2a. Site A and Site C have the same bounds and Site C and Site D have the same bounds. The two piecewise average is used to build the final model.

Another approach allows for limited data exchange across clients to expand these bounds. In Fig 2b., Client A and Client B exchange limited data. But this relaxes the weakens the "federated" learning since no data exchange should occur in federated learning and such a data exchange may have implications regrading to the specialized compliance requirements.

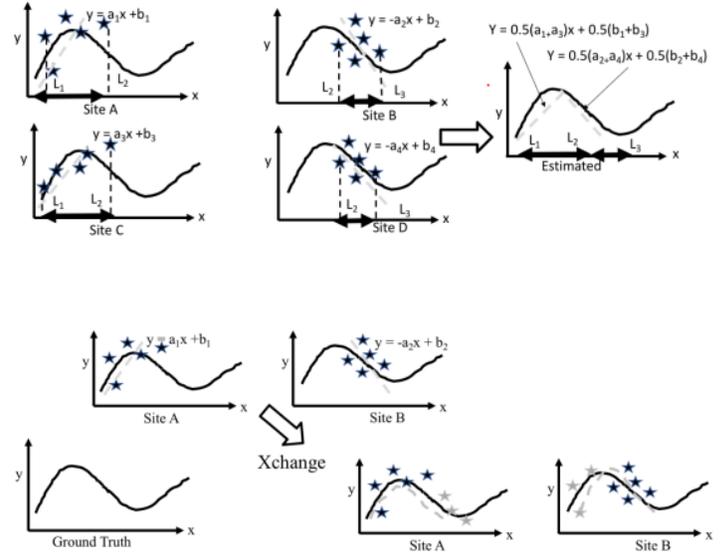

**Fig 2: a) Bound aware averaging b) Limited data exchange method from the paper "Approaches to address the data skew problem in federated learning" [16].**

Depending on the cases presented above, data reconstruction can also be performed on the client devices using sampling techniques, for example, using data oversampling techniques like Adaptive Synthetic Sampling (ADASYN) and Synthetic Minority Oversampling (SMOTE). Also, specialized machine learning models like Generative Adversarial Networks (GANs) can be used to synthesize the data without any data exchange across the client devices [17].

## 4. APPROACH
**Research Methodology:** Experiments

Conducting practical experiments for this study with clients and servers is difficult because of the time it takes to implement and set up these systems. An alternative approach is to simulate the client devices and conduct experiments.

As discussed in the previous section, using TFF it is possible to simulate the federated learning algorithms with their respective model and data. The simulation datasets are quite representative of the real-world datasets and hence the result from these simulations is expected to be identical to the real-world scenario.

TFF is also currently enabling high performance simulations to get faster results in experimentations but the work is in progress. There are a variety of Machine learning tasks like Computer Vision and Natural Language processing and TFF provides simulation datasets for all kinds of ML tasks. But for this project, the classic EMNIST Handwritten Images dataset for the image recognition task was used to characterize the issues and conduct experiments.

**Table 1. EMNIST Simulation Dataset**

| Type | No. of examples |
|---|---|
| Train | 341,873 |
| Test | 40,832 |

The EMNIST dataset consists of both handwritten images and characters. For simplicity of conducting experiments, only the digits are considered since the underlying task definition remains the same. The simulation dataset consists of 3383 users (client devices) with each client having data for the 10 classes. (digits 0 - 9).

Since Machine Learning experiments require significant computing power, the experiments were conducted on Google Colab, a cloud based interactive Python notebook that offers GPU and TPU runtimes.

## 5. IMPLEMENTATION

The TensorFlow Federated (TFF) is a strongly typed programming environment. It provides sets of APIs at two different levels of abstractions. A set of lower-level Federated Core (FC) APIs and a set of higher-level Federated Learning (FL) APIs.

The FL APIs is built from FC and allows researchers to experiment and evaluate the performance, implement optimizations for *existing* federated learning algorithms.

The FC APIs provide a strongly typed functional programming environment for experimenting with *novel* Federated Learning Algorithms that combines TensorFlow with distributed communication operators.

For conducting experiments in this project, the following interfaces were used, and it is necessary to understand them at a higher level to understand the implementation of the federated learning process in the TFF framework.

tff.simulation.datasets interface – To access the simulation datasets.

tff.learning.Model interface – Exposes methods to stamp the model's forward pass and metadata properties.

tff.learning.from_keras_model – When the underlying ML model is built on Keras, this can be directly used to wrap the model.

tff.learning.build_federated_averaging_process interface – To build the Federated Averaging algorithm.

client_optimizer_fn – To create the optimizer that performs the local model update on each client.

server_optimizer_fn – To create the optimizer performs the averaged update to the global model.

Keras is a higher-level API for TensorFlow and the development team of TFF encourages using Keras to build the initial model. Hence in this project, the models are created using Keras and once created, the model is wrapped to be an instance of tff.learning.from_keras_model, passing the input client data and model definition as arguments.

To implement the Federated average process, the tff.learning.build_federated_averaging_process interface is invoked and the model definition is passed as a constructor along with the client_optimizer_fn and server_optimizer_fn.

TFF will construct a pair of federated computations and pack them into tff.templates.IterativeProcess iterator in which the computations are available as pair of *Initialize* and *next*. These computations are used to implement the Federated averaging algorithm.

To execute these computations, regular function calls are used. This however hampers the performance of TFF and the support for high performance simulations is still under development.

The *Initialize* computation is invoked as a function that takes no argument and returns the state of the federated averaging process on the server. The state of the server consists of the model and the server's optimizer state.

The *next* computation is also invoked as a function and it represents a single round of the federated averaging process ie., from pushing the server state and model parameters to the clients, training the models locally on the client devices, aggregating and averaging the model updates to produce a new updated model in the server.

To simulate the case in which not the same client devices participate in each round but only a subset, a random sampling of client devices is performed in each round. This however slows down the convergence of the learning algorithm.

The federated training metrics will indicate if the model training is progressing i.e.., lowering of loss values and increasing accuracy in each round.

TFF also provides support to use TensorBoard, a popular visualization tool used along with TensorFlow. For each round, the training metrics are logged which can be later visualized using TensorBoard.

The initial model can also be created using the tff.learning.Model, which is a lower-level API that allows for maximum customization of the models and parameters without changing the underlying federated averaging process.

One caveat of the training metrics is that it cannot provide information if the local models in clients are overfitting or not, it can only be interpreted as progression in model training. To evaluate the trained model on federated data, TFF provides another federated computation tff.learning.build_federated_evaluation which is again invoked as a function and takes the model constructor as the input argument.

Unlike tff.learning.build_federated_averaging_process, the tff.learning.build_federated_evaluation does not modify the server states since it does not perform optimizations.

During test time, if the test data is centralized then TFF also provides support through tf.keras.models.Model.evaluate which applies the weights from federated training to a Keras model which can then be used on the centralized test data to evaluate its performance.

# 6. EXPERIMENTATION

## 6.1 Heterogeneity of the simulation dataset

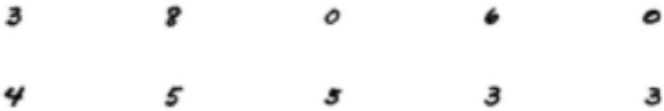

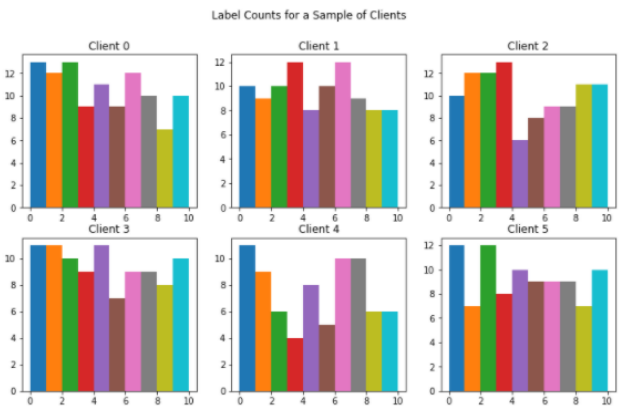

**Fig 3: a) Data from a single client device. b) Distribution of classes across the client devices.**

Each client device has a particular "pattern" of handwriting as seen in Fig3. a, since it is a representation of data in a single client device. In the simulation datasets by default, IIDness is achieved since the number of samples in each client device is approximately equal and every client has enough data in all the classes.

On a note of technicality, the simulation dataset consists of the handwriting of one user per client device. But in reality, multiple users may be using the same devices adding to the statistical heterogeneity challenges. This project assumes that the distribution of each modality in a client device is the same ie., one user per client device.

## 6.2 Experiment 1: Characterization of Communication overhead

In the first experiment, a lossy compression technique is applied for the model broadcasts and gradient aggregations to find if compressing the models as discussed in 3.4. reduces the model's performance significantly i.e.., The goal is to characterize the overhead issue and find the effects of compression on model's performance (accuracy).

This experiment is carried out in Google Colab with a TPU. The Federated averaging process is constructed as discussed in section 5.

The lossy compression algorithm to be applied on the broadcasted and aggregated data is built using the tensor_encoding API.

The encoder functions are first defined as an instance of the core interfaces te.core.SimpleEncoder and te.core.GatherEncoder.

TFF provides APIs for the tff.learning.build_federated_averaging_process to consume the converted encoder function.

tff.learning.framework.build_encoded_broadcast_from_model– This interface creates a broadcast process that takes the model function and broadcast encoder function as inputs.

tff.learning.framework.build_encoded_mean_from_model – This interface creates the aggregate process that takes the model function and mean encoder function as inputs.

The above two processes are passed as arguments to the tff.learning.build_federated_averaging_process interface along with the model constructor, client and server optimizers as discussed in the previous section.

The model used is a Convolutional Neural Network (CNN, as implemented in McMahan's work [10]. Creating a realistic complex model for the task provides a realistic estimate for the communication costs.

The compression is not applied to the entire model. Each variable of the model is compressed independently since the effectiveness of applying to a smaller variable is relatively small. Variables that have more than 10000 elements are chosen and a uniform quantization to 8 bits (256 buckets) is applied. Ten clients were used in each round to train a model for 1 epoch locally. The clients are chosen by randomly sampling a subset of clients in each round for 250 rounds.

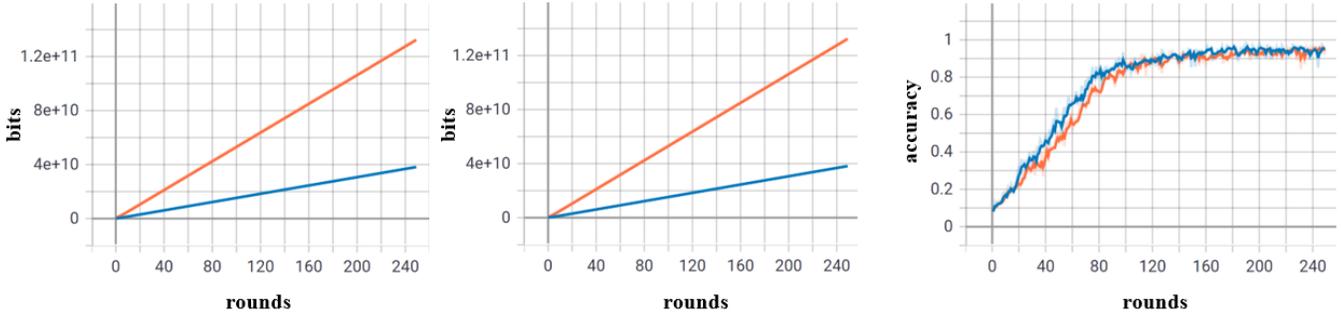

**Fig 4: a) Cumulative aggregated bits b) Cumulative broadcasted bits c) Accuracy of the trained model**

The orange curves represent before compression and the blue curve represents after compression in a, b and c.

The experimentation result is shown in Fig 4., and by using a lossy compression technique the cumulative broadcasted bits i.e.., the bits sent from the central server to clients, and the cumulative aggregated bits i.e.., the bits sent from clients to central server, are greatly reduced from 17 GB before compression to 5 GB after compression, and the compressed processes' final model performance (accuracy) is on par with the final model's performance from the non-compressed federated process after carefully tuning the model and optimizer's hyperparameters (learning rate, batch size, number of rounds, number of client devices)

The accuracy is 93% before compression and approximately 92.7% after compression after compression. The noise in the accuracy curves is because a random subset of clients is chosen to simulate that client's participating in each round are stateless, i.e., client participating in one round may not participate in the next round due to one of the reasons discussed previously.

## 6.3 Experiment 2: Characterization of non IID-ness

In the second experiment, the goal is to characterize that non IID-ness affects the model performance. There are many notions to generate non IID, but this project will focus on the labels non IID, in which each client has only one class of the data as shown in Fig 5., where in each plot, x-axis is the classes [0 - 9] and y-axis is the number of samples in that class.

Generating non IID simulation datasets is not at the core of the TFF. But there are some helper functions to create label non IID such as tff.simulation.datasets.build_single_label_dataset.

This experiment was performed on 10 client devices with a batch size of 20. Each client had only one of the classes with local model training for 5 epochs for 20 rounds.

Since the goal is to check the effect of non IID-ness the communication costs associated with federated learning process can be lowered by using a much simpler model thereby aiding in faster simulation results.

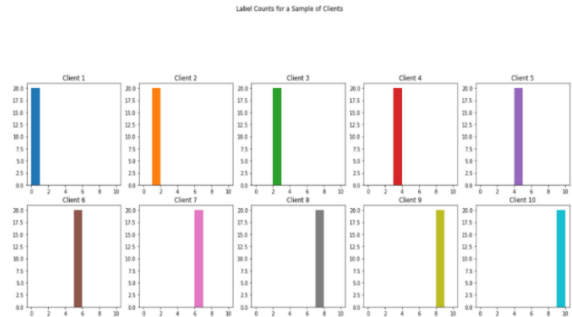

**Fig 5: Each client device has only a single label – Label Non IID**

The model consists of a single hidden layer of 10 units followed by a softmax classifier. This lowered costs and faster experimentation results allow increasing the number of rounds and client devices used in the experiment.

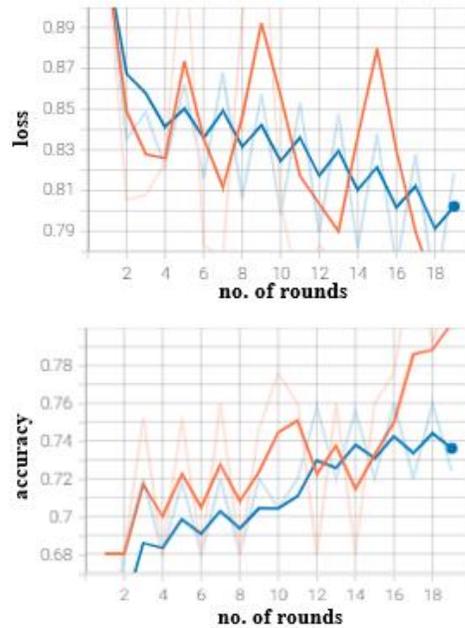

**Fig 6: a) Loss curves b) Accuracy curves**

In Fig 6., the orange curves represent the IID Dataset and the blue curve represents the non IID dataset, and it is evident that the label non IID-ness causes poor model performance. The noise in the accuracy curves is again due to the fact that random samples of clients were chosen in each round. The accuracy of IID model is 80 % while the accuracy of the non IID model is 73%.

The oversampling technique seen in section 3.6 might not work in this extreme case of each client having only one label. The alternative solution to overcome this case of non IID-ness would be to allow limited data exchange across the client devices.

## 7. CONCLUSION

Federated Learning is a privacy focused machine learning technique that borrows many ideas from distributed machine learning. However, there are still many open challenges in FL that are not yet addressed in this project. This is an active research area and the experimental frameworks although eases conducting experiments, the programmers still need to learn about various APIs.

This project focused specifically on the communication bottleneck and the non IID-ness issue in FL, and these issues were characterized in a baseline model. The compression technique seems to be a promising approach to minimize the communication costs without affecting the final model's performance significantly.

## 8. FUTURE WORK

Experiment 1 used a lossy compression technique. Custom implementation of various compression algorithms is possible, and more research is still needed for choosing an appropriate compression technique for a specific machine learning task. It is to be investigated if the same quantization technique will work for natural language processing tasks by reducing communication cost without reducing the model's performance.

Experiment 2 proves that the non IID-ness does impact the mode's performance. The various techniques for correcting the non IID-ness (discussed in section 3.6.,) could be implemented in the future. Also, this project only focused on the label non IID-ness and there are opportunities to explore other notions of Non IID-ness.

This project will also be carried in a peer-to-peer machine learning setting. The communication bottleneck and the effects of non IID-ness on final model are still concerns that need to be addressed in a peer-to-peer learning environment, and the decentralized orchestration in such systems adds to the complexity.

Finally, the discussions in this project are solely based on model based federated learning. There is a new approach to federated learning called data centric federated. In a data centric federated learning, instead of the model, the data is hosted on a peer-to-peer network.

Any data scientist can request access to the data and perform experiments. Frameworks such as PyGrid may be used by organizations that would like to host their data [22] rather than the model, allowing data science experts to collaboratively train models.

## 10. ACKNOWLEDGEMENT

The author thanks Prof. Ashvin Goel, University of Toronto, and the TensorFlow Federated Development team for their continuous support throughout this project.